\documentclass[conference]{IEEEtran}
\usepackage{times}

\usepackage[numbers]{natbib}
\usepackage{multicol}
\usepackage[bookmarks=true]{hyperref}

\usepackage{amsmath,amsfonts,bm}

\def\eqref#1{equation~\ref{#1}}
\def\1{\bm{1}}

\DeclareMathAlphabet{\mathsfit}{\encodingdefault}{\sfdefault}{m}{sl}
\SetMathAlphabet{\mathsfit}{bold}{\encodingdefault}{\sfdefault}{bx}{n}

\newcommand{\KL}{D_{\mathrm{KL}}}

\DeclareMathOperator*{\argmax}{arg\,max}
\DeclareMathOperator*{\argmin}{arg\,min}

\DeclareMathOperator*{\Ebig}{\mathbb{E}}

\usepackage{hyperref}
\usepackage{url}

\usepackage{bbm}
\usepackage{graphicx}
\usepackage{caption}
\usepackage{subcaption}

\usepackage{amsmath}
\usepackage{xcolor}
\usepackage{listings}

\newcommand{\rlive}{r_\mathrm{live}}
\newcommand{\rdone}{r_\mathrm{done}}

\title{Learning Sparse Control Tasks from Pixels by Latent Nearest-Neighbor-Guided Explorations}

\author{Ruihan Zhao$^1$ \; Ufuk Topcu$^1$ \; Sandeep Chinchali$^1$ \; Mariano Phielipp$^2$\\
$^1$The University of Texas at Austin \; $^2$Intel AI Lab}

\begin{document}

\maketitle
\IEEEpeerreviewmaketitle

\begin{abstract}
Recent progress in deep reinforcement learning (RL) and computer vision enables artificial agents to solve complex tasks, including locomotion, manipulation and video games from high-dimensional pixel observations. However, domain specific reward functions are often engineered to provide sufficient learning signals, requiring expert knowledge. While it is possible to train vision-based RL agents using only sparse rewards, additional challenges in exploration arise. We present a novel and efficient method to solve sparse-reward robot manipulation tasks from only image observations by utilizing a few demonstrations. First, we learn an embedded neural dynamics model from demonstration transitions and further fine-tune it with the replay buffer. Next, we reward the agents for staying close to the demonstrated trajectories using a distance metric defined in the embedding space. Finally, we use an off-policy, model-free vision RL algorithm to update the control policies. Our method achieves state-of-the-art sample efficiency in simulation and enables efficient training of a real Franka Emika Panda manipulator. Code will be made available online. \footnote{Project website: \href{https://philipzrh.com/latent-nn}{https://philipzrh.com/latent-nn}}
\end{abstract}

\section{Introduction}
Recently there has been tremendous success in using deep reinforcement learning for various tasks involving decision-making. Deep RL is, in principle, a versatile approach that can directly learn from interaction data, often without an explicit, hand-coded dynamics model.
By interacting with environments, RL agents can learn optimal policies from either dense or sparse environment feedback. A rich collection of state-of-the-art approaches allows the learning of policies for both discrete actions and continuous action spaces, while taking in either low-dimensional state vectors or high-dimensional sensor readings \cite{sac, dqn, ppo, trpo}.

However, it remains challenging to apply Deep RL to real-life domains, including real-hardware robot learning. One major challenge comes from the need to reliably track the complete system state \cite{dexterous}. To mitigate this challenge, learning a policy that directly maps camera readings to optimal actions requires much less engineering effort. Recently, by incorporating ideas from the computer vision domain, including data augmentation and self-supervised learning, state-of-the-art RL algorithms can learn control policies purely from image observations, with high sample efficiency \cite{dreamerv2, curl, rad, drq}. On the other hand, it is also difficult to assign credit to the learning agent in a scalable way. Reward engineering utilizes domain-specific knowledge to provide better training signals: popular simulated environments provide optional dense reward functions based on heuristics \cite{gym, robosuite2020}, but might rely on state information not readily available in the real physical system. Thus, there has been a lot of effort to help RL agents explore more effectively in environments with sparse or no rewards \cite{diayn, vic}. For example, a sparse reward setting might be a manipulation task where a robot receives a positive reward only at task completion (e.g., for successfully placing an item on a shelf) and a zero reward at all other time steps. 

\begin{figure}[t]
    \centering
    \includegraphics[width=0.95\linewidth]{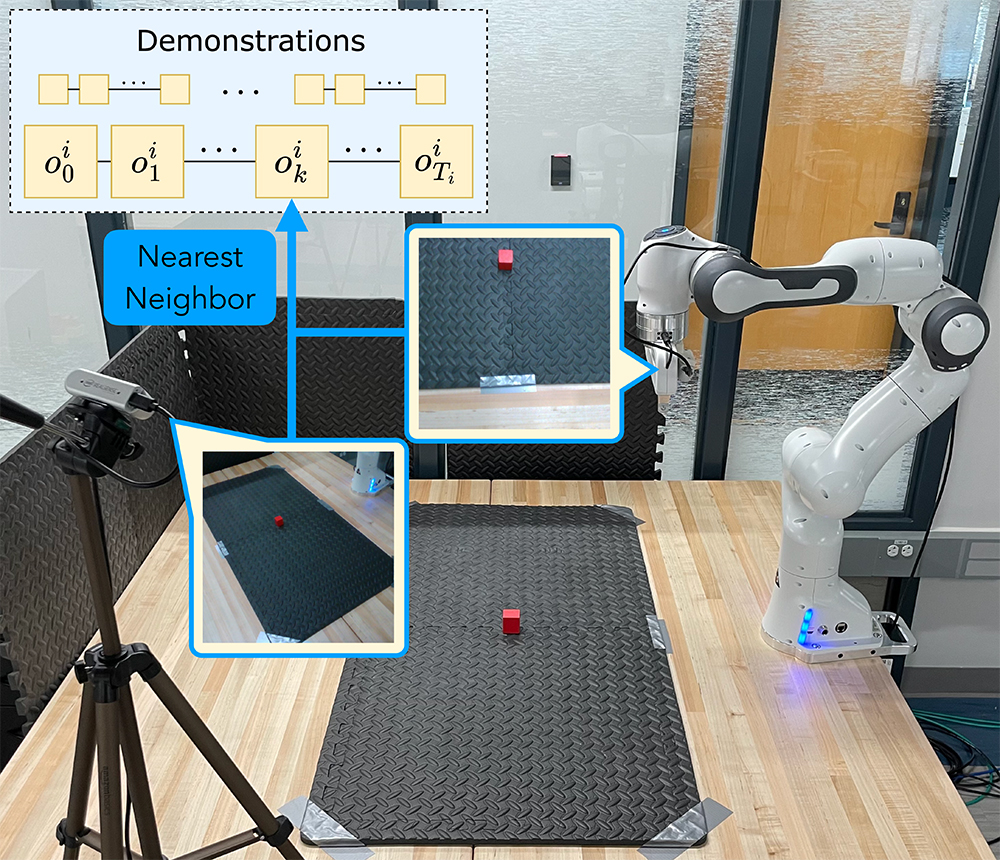}
    \caption{\small We enable efficient model-free reinforcement learning on a real Franka Emika Panda robot with 7 degrees of freedom from RGB image observations, sparse rewards, and only a few demonstrations. This is achieved by learning a distance metric in an embedding space, and rewarding the agent for staying close to the demonstrations.}
    \label{fig:franka}
\end{figure}

\begin{figure*}
    \centering
    \includegraphics[width=\linewidth]{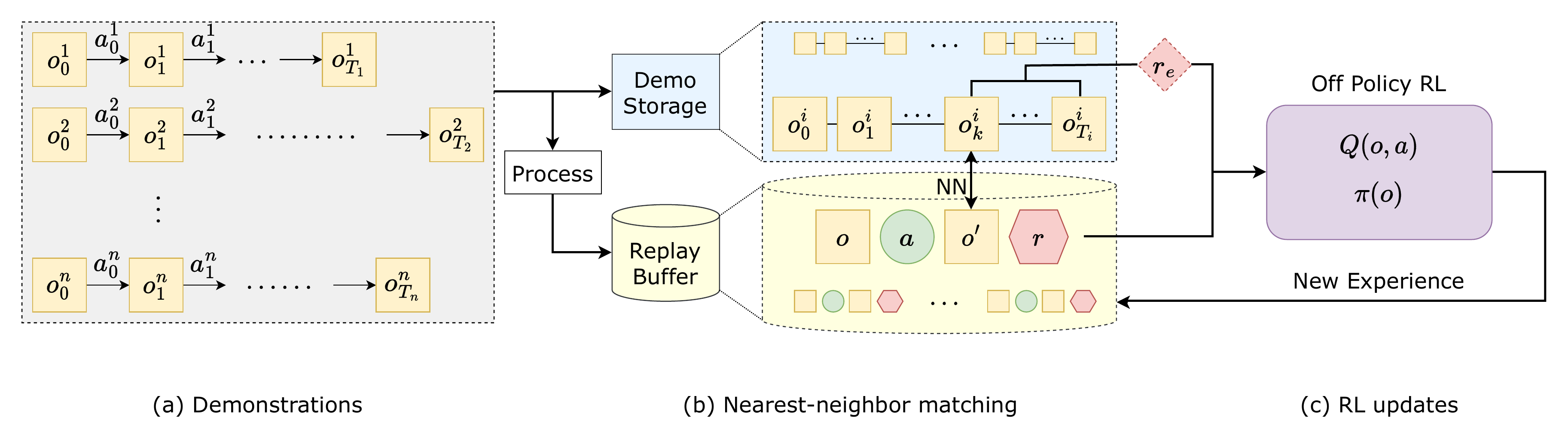}
    \caption{\small Our approach augments the sparse task reward with a dense exploration reward to enable efficient training of vision-based robotic manipulation policies using Deep Reinforcement Learning. (\textbf{Left}) We have a set of variable length demonstrations $i$, each consisting of observations $o^i_t$, actions $a^i_t$ etc. (\textbf{Middle}) During learning, we update our policy with a mixture of stored demonstrations (blue) and examples from the replay buffer (yellow). 
    (\textbf{Right}) Our RL agent maps a sensor observation $o$ to an action $a$. It finds the closest demonstration to infer a dense reward $r'$ based on how close we are to the goal.}
    \label{fig:method}
\end{figure*}

In this work, we propose a novel approach to tackle the exploration challenge in image-based control tasks with sparse rewards. We build on the idea of using demonstrations to help with exploration and draw inspiration from reward shaping, which has been explored in prior works \cite{modem, coder}. The main insight is that each step in the demonstration can be considered a checkpoint, and the agent should be credited for reaching a similar state (see Fig. \ref{fig:method}). Our main contributions are two-fold. First, we define a distance metric between pairs of image observations from the task domain by learning an embedding from images to a lower-dimensional latent space, making it possible to quantify how close the agent is from a demonstration state. Second, we design a systematic way to utilize the learned distance metric to provide much denser reward signals in sparse-reward tasks.
Our approach is independent of the underlying RL algorithm. In this work, we implement our approach on top of the state-of-the-art off-policy algorithm SAC \cite{sac}. Simulation results show that our method significantly improves sample efficiency on long-horizon, sparse-reward tasks compared to state-of-the-art model-based and model-free methods. In the real world, our method enables the learning of manipulation tasks from scratch with only a few hours of training and a handful of demonstrations.

\section{Related Work}
During the training of an RL agent, exploring efficiently is challenging especially in sparse-reward environments -- agents need to interact with the rewards sufficiently to obtain meaningful learning signals. One popular approach to provide denser rewards is by incorporating human and domain knowledge. For example, reward shaping is usually used in long-horizon control tasks by adding intermediate rewards at important checkpoint locations \cite{robosuite2020}, or a dense reward is derived from physical understanding of the underlying system \cite{reward_shaping}. In the domain of real robot manipulation, a recent work by \citet{reward-sketching} takes advantage of imperfect dense reward signals provided by human operators. During training, engineers annotate episodes of robot experience to help train a reward model.

Aside from efforts to provide dense rewards, novel algorithms are proposed to improve the agents' exploration behavior. For instance, maximum entropy RL balances the agent's exploration and exploitation behavior by encouraging exploration as part of the learning objective \citep{sac}. Hierarchical RL and intrinsically motivated skill learning are also potentially remedies. For example, \citet{diayn} proposed to learn a set of diverse low level skills, which can then be used as action primitives for a high-level RL agent to explore larger areas of the state space. \citet{curiosity} propose to utilize an intrinsic reward to model curiosity. In this approach, the agent learns an internal model of the world, and is awarded for visiting states that are not captured well by its model.

It has also been shown that warm-starting and demonstrations are helpful in sparse-reward tasks. \citet{qt-opt} learn a closed-loop grasping policy purely from real robot data, with only binary reward signals. To assist with initial exploitation, the method employs a warm-start technique by using a scripted policy as the starting point. \citet{sparse_rewards} propose an algorithm that alternates between updating the RL policy from the environment and pulling it towards demonstration behaviors using an information theoretic approach. \citet{coder} employs a contrastive learning objective on demonstration trajectories to pre-train the image feature extractors, while also retaining them for RL updates. \citet{reward-sketching} provide demonstrations through tele-operation as part of the training loop.

Finally, model-based RL methods aim to improve training sample efficiency by learning a transition model of the environment from past experience, and generate new training samples for RL updates. \citet{dreamerv2} learns an accurate discrete world model from high dimensional image inputs for Atari games which enables the training of an RL agent that achieve human-level performance. \citet{modem} combines model-based learning with demonstrations by over-sampling demonstrated data to form a behavior prior.

The key innovation of our work compared to prior art is that we introduce a systematic pipeline for learning sparse reward tasks from high-dimensional observations using model-free reinforcement learning. As shown in Fig. \ref{fig:method}, our innovation comes from (a) matching to similar demonstrations using a locally linear latent model, (b) using data augmentation to improve the robustness of state matching, and (c) inferring a dense reward based on promising states in the demonstrations.

\section{Background}

\subsection{Deep Reinforcement Learning}
In this work, we adopt a typical deep reinforcement learning setting, where the underlying dynamics of the environment is modeled as a Markov Decision Process (MDP). The MDP is defined by the tuple $(\mathcal{S}, \mathcal{A}, p, r, \gamma)$ where $\mathcal{S}$ is the state space and $\mathcal{A}$ is the action space. $p$ models the state transition probability, where $p(s' \,|\, s, a)$ is the chance of landing at state $s'$ after taking action $a$ from state $s$. The reward function $r(s, a, s')$ indicates the reward given to the agent for taking this transition. Finally, $\gamma$ is the discounting factor. An RL algorithm aims to find the optimal policy $\pi^*(a \, | \, s)$ that maximizes the agent's cumulative reward in expectation.
\begin{equation}
    \label{eqn:rl}
    \pi^* = \argmax_{\pi(a \, | \, s)} \mathbbm{E}_\pi\bigg[\sum_{t = 0}^\infty \gamma^t \cdot r(s_t, a_t, s_{t+1})\bigg].
\end{equation}
To reflect the use of image observations, we denote $\mathcal{O}$ as the observation space, and $o_t$ as the input images at time $t$. For the tasks that we solve, we assume that the necessary environment states are captured within our image observations.

\subsection{Problem Setting}
This work tackles the challenges that come with vision-based RL in long-horizon, sparse reward tasks, where the agent only receives a positive reward $\rdone$ at task completion, while getting negative rewards $\rlive$ everywhere else. The interpretation of such a reward function is that the agent gets a high reward only at task completion, but is also penalized for the length of the trajectory. Formally, letting $\mathcal{G}$ denote the set of goal states, we define the reward function as follows:
\begin{equation}
    \label{eqn:reward}
    r(s, a, s') =
    \begin{cases}
        \rdone > 0 & s' \in \mathcal{G}\\
        \rlive < 0& \text{otherwise}.
    \end{cases}
\end{equation}
The sparse reward function reduces the need for expert knowledge or human intervention, making it much easier to implement in a real-world environment. However, this makes exploration hard: the robot gets meaningful signal only when it reaches a goal, which is unlikely in long-horizon complex control tasks. We now propose a method to address the challenge of vision-based, sparse reward robotic tasks.

\section{Method}
We propose a training pipeline centered around a few demonstrations to tackle the exploration challenge with sparse-reward learning. The core idea is to provide additional dense rewards at states close to the demonstrations. To correctly measure the distance between states from their image observations, we find a structured embedding space by learning a latent dynamics model as an auxiliary task. In addition, we employ value clipping to take advantage of the sparse reward structure, and importance sampling to prioritize learning signals from demonstrations.

\subsection{Off-policy Reinforcement Learning with Demonstrations}
\label{subsec:rl-demo}
We provide a set $\mathcal{D}$ of demonstrations consisting of $n$ successful trajectories of observations and actions. Each demonstration trajectory $i$ may be of different length $T_i$, but must terminate in the goal set $\mathcal{G}$. We assume the demonstrations to come from a human operator or a heuristic controller, and thus, can be imperfect in nature. We formalize the mathematical notations below, where $\mathcal{T}^i$ denotes the trajectory for demonstration $i$. $s_t$ is the underlying true state of the robot and $o_t$ is the high-dimensional observation received at time $t$, such as a pair of camera images. $a_t$ is the action taken by the demonstrator. Note that we don't have access to the ground truth state $s_t$ in the demonstrations, as shown below.
\begin{align}
    \label{eqn:demo}
    \mathcal{D} &= \{\mathcal{T}^1 , \mathcal{T}^2 \cdots \mathcal{T}^n \} \nonumber\\
    \mathcal{T}^i &= (o_0^i , a_0^i , o_1^i , a_1^i , \cdots o_{T_i - 1}^i, a_{T_i - 1}^i , o_{T_i}^i)\\
    \forall i &, s_{T_i}^i \in \mathcal{G} \nonumber .
\end{align}
The demonstrations are kept in two forms. First, the trajectories are stored to help keep track of the steps-to-success for each state. This information is used to properly discount the exploration reward, as will be discussed later. Next, they are also sliced into experience tuples $(o, a, o', r, d)$ and placed in the replay buffer $\mathcal{B}$ of our off-policy, model-free reinforcement learning algorithm. Here, the notation $o'$ means the next observation after observation $o$. $r$ is the sparse reward signal as defined in Equation \ref{eqn:reward}, and $d$ is a Boolean variable indicating episode termination. These experience tuples are essential to propagate the reward signals into the value function and policy. As the replay buffer fills from new experience, we make sure the demonstration transitions are never ejected.

\subsection{Augmentation-invariant Distance Metric}
Our next goal is to calculate the distance between the robot's current observation during learning and the closest observation in the demonstration set. This will allow the robot to infer how close it is to achieving the sparse reward.
Computing the distance between two states from their respective image observations is challenging: two drastically different states might only differ by a few pixels. Conversely, the same underlying state might appear very different in two camera observations due to task-irrelevant background features.
Thus, we propose to embed the image observations into a low-dimensional latent space to obtain a viable distance metric. Inspired by \citet{e2c}, we utilize a Variational Auto-encoder (VAE) \cite{vae} and a locally-linear dynamics model to regularize the structure of the latent space. The locally-linear dynamics model captures our goal for the latent space to be temporally consistent since we have a multi-step control task.

Meanwhile, it has been shown that data augmentation is essential for efficient and robust image-based Reinforcement Learning \cite{rad, drq}. To make the distance metric invariant to data augmentations, we train the VAE and latent dynamics model to encode augmented observations, while predicting the unchanged images. We use $f(\cdot)$ to denote the random augmentation function. The trainable components include the CNN encoder $E_{\phi}$, the decoder $D_{\theta}$, and the latent transition model $M_{\psi}$. 

Our model is trained by optimizing a variational evidence lower bound (ELBO) objective, using transition tuples $(o, a, o')_i$ sampled from the replay buffer. We assume a unit Gaussian prior $p(\mathbf{z})$ in the latent space. Additionally, the encoded distributions $q(\mathbf{z} \,|\, o)$ and decoded distribution $p(\mathbf{o} \,|\, z)$ are also modeled as Gaussian distributions, which is a good fit for predicting colored images. First, the data-augmented observations are encoded into their latent distributions whose mean $\mu$ and diagonal covariance matrix $\Sigma$ are predicted by the encoder network $E_{\phi}$:
\begin{align}
    \begin{split}
    z \sim q_\phi(\mathbf{z} \,|\, f(o)) = \mathcal{N}(\mu, \Sigma) &, \; (\mu, \Sigma) = E_{\phi}(f(o)) \\
    z' \sim q_\phi(\mathbf{z'} \,|\, f(o')) = \mathcal{N}(\mu', \Sigma') &, \; (\mu', \Sigma') = E_{\phi}(f(o')).
    \end{split}
\end{align}

The one-step dynamics model in the latent space is locally linear in the state and action, whose parameters (matrices $A$, $B$ and offset $c$) depend on the starting state, as predicted by the latent transition model $M_{\psi}$. Here $z$ is the latent state representation predicted by the encoder. Prior work shows that a latent linear dynamics model is tractable to learn, but provides modeling flexibility through local linearity \cite{e2c}. The latent dynamics are given by:
\begin{equation}
    \hat{z'} = Az + Ba + c, \hspace{0.5cm} (A, B, c) = M_{\psi}(z).
\end{equation}

The linear transition model allows the prediction of the next step latent distribution using the starting distribution and action as follows:

\begin{equation}
    q_{\psi}(\mathbf{\hat{z'}} \,|\, z, a) = \mathcal{N}(\hat{\mu'}, \hat{\Sigma'}),
\end{equation}
where
\begin{align}
    \begin{split}
        \hat{\mu'} &= A \mu + B a + c \\
        \hat{\Sigma'} &= A \Sigma A^T.
    \end{split}
\end{align}

The encoder $E_{\phi}$, decoder $D_{\theta}$, and transition model $M_{\psi}$ are updated jointly using a combined loss function which comprises of 3 objectives. First, we want the sampled starting latent state $z$ to be reconstructed back to the original observation $o$. Similarly, as we pass the sample through the dynamics model, the resulting latent state prediction $\hat{z'}$ should reconstruct back to $o'$. Finally, to ensure the latent dynamics model is consistent over multiple steps, we want the predicted distribution $q_\psi(\mathbf{\hat{z'}} | f(o), a)$ and encoded distribution $q_\phi(\mathbf{z'} | f(o'))$ to be similar. Formally, we write the overall training objective $\mathcal{L}$ as follows, where $\gamma$ is a hyper-parameter for weighing the two loss terms:
\begin{align}
    \begin{split}
    \mathcal{L}_\mathrm{ELBO} &= \Ebig_{z \sim q_\phi, \, \hat{z'} \sim q_{\psi}} \bigg[ -\log p(o | z) - \log p(o' | \hat{z'}) \bigg] \\
    &+ \KL \bigg(q_\phi(\mathbf{z} \,|\, f(o)) \;\bigg\Vert\; p(\mathbf{z})\bigg)
    \end{split}\\
    \mathcal{L}_{\mathrm{dynamics}} &= \Ebig_{z \sim q_\phi} \bigg[ \KL \bigg(  q_\psi(\mathbf{\hat{z'}} \,|\, z, a) \;\bigg\Vert\; q_\phi(\mathbf{z'} \,|\, f(o'))\bigg) \bigg]\\
    \mathcal{L} &= \Ebig_{(o, a, o') \in \mathcal{B}} \bigg[ \mathcal{L}_\mathrm{ELBO} + \lambda \mathcal{L}_\mathrm{dynamics} \bigg].
\end{align}

In essence, we want to minimize the reconstruction error for the VAE using the ELBO objective (term 1), but also want to minimize the dynamics prediction error in the latent space (term 2). Crucially, given the learned encoder, we can now define a task-relevant distance metric between high-dimensional observations $o$. Specifically, given two augmented observations $f(o_1)$ and $f(o_2)$, we define the Augmentation-invariant Distance Metric (ADM) to be the $L^2$ distance between the augmented and \textit{encoded} states:
\begin{equation}
    \label{eqn:distance_metric}
    d(f(o_1), f(o_2)) := ||E_{\phi}(f(o_1)) - E_{\phi}(f(o_2))||_2.
\end{equation}
We now describe how to use the above Augmentation-invariant Distance Metric to find the closest demonstration to the robot's current observation.

\subsection{Demonstration-Guided Exploration}
To encourage the agent to explore more efficiently, we propose a reward-engineering approach to credit the agent for staying close to demonstrations.
Given an experience tuple $(o, a, o', r, d)$, we assign an additional exploration reward $r_e$ if $o'$ is sufficiently close to an observation in our demonstration, up to the distance threshold $\epsilon$, which is dynamically computed. We now formally define the exploration reward $r_e$ and the distance threshold $\epsilon$.

First, we compute the distance threshold $\epsilon$, which is defined as the average distance between consecutive observations within the demonstrations. In this way $\epsilon$ approximates the distance of one environment step. As we get new experience from learning, we re-compute the distance threshold $\epsilon$. Re-computation is necessary because the encoder $E_{\phi}$, decoder $D_{\theta}$, and dynamics model parameters $M_{\psi}$ are constantly updated with the latest experience to ensure our distance metric is not overfitted to only the demonstration data. The definition of $\epsilon$ is formulated as follows.

\begin{equation}
    \label{eqn:eps}
    \epsilon := \Ebig_{i, t}\big[ d(o_t^i, o_{t+1}^i) \big] , \hspace{0.5cm} o_t^i, o_{t+1}^i \in \mathcal{D}
\end{equation}

Next, we find the trajectory index $i$ and time step $t$ of the nearest demonstration observation to our observation $o'$ using the distance function $d$:

\begin{equation}
    \label{eqn:dge}
    i^*, t^* = \argmin_{i, t}d(o', o_t^i), \hspace{0.5cm} o_t^i \in \mathcal{D}.
\end{equation}

Our goal is to create a dense reward to encourage the agent to reach promising states which led to task success in the demonstrations. Thus, we add an additional exploration reward $r_e$ if we are close to a demonstration state and modulate it by how far we are expected to be from success. In our experiments, $r_e = 1$. First, we discount the exploration reward $r_e$ by powers of $\alpha$ depending on the number of steps until success; the exploration reward is greater when closer to the goal. The discounting factor $\alpha$ is a hyper-parameter chosen independently from the RL discounting factor $\gamma$. The extra reward is added to the environment reward $r$, and the resulting sum $r_{\mathrm{dense}}$ is used to update the RL agent. As shown in the equation below, we focus on the demonstration with index $i^*$, where we find our nearest neighbor observation. $T_{i^*}$ is the length of this demonstration trajectory, as introduced in Section \ref{subsec:rl-demo}. Note that we don't give an exploration reward if the state is already a successful state. Thus, our final reward becomes: 

\begin{equation}
    \label{eqn:demo_reward}
    r_{\mathrm{dense}} =
    \begin{cases}
        r + \alpha^{T_{i^*} - t^*} r_e & [d(o', o_{t^*}^{i^*}) \leq \epsilon] \land [o' \notin \mathcal{G}]\\
        r & \text{otherwise}.
    \end{cases}
\end{equation}

We observe that in the case where $o'$ finds its nearest neighbor from one of the successful terminal states, we have $t^* = T_{i^*}$. Then, $o'$ is awarded the full exploration reward $r_e$ along with the nominal reward $r$. Whereas, when $o'$ is close to one of the earlier steps in a demonstration, $r_e$ is heavily discounted. Finally, if we are very far from any demonstration observation (relative to the distance threshold $\epsilon$), we simply pass the RL agent the nominal reward $r$ (case 2 in Eq. \ref{eqn:demo_reward}). Or, if we are at the goal, we get the goal reward in case 2 of Eq. \ref{eqn:demo_reward}.
Given this new reward $r_\mathrm{dense}$, we can train using any off-the-shelf RL algorithm, which leads to the versatility of our approach.

\subsection{Importance Sampling and Value Clipping}
To further improve training efficiency, we take advantage of the reward and demonstration structures by implementing importance sampling and $q$-value clipping. Both importance sampling and $q$-value clipping are standard tools in RL to improve learning efficiency, but in our case we have a rigorous derivation of the specific bound we choose.

We perform importance sampling when choosing transitions from the replay buffer for updating the RL agent. When sampling a batch of $b$ transitions, we prioritize demonstration transitions so that they constitute at least $p_d$ fraction of the batch. This technique allows the task reward $\rdone$ to propagate quickly into the values of promising states and for the agent to effectively learn from demonstrations. The batch size $b$ and fraction $p_d$ are hyper-parameters.

To further take advantage of the sparse reward structure and stabilize training, we scale the three reward sources $\rdone$, $\rlive$ and $r_e$ so that the resulting $q$-value is bounded, allowing us to clip the value target when updating $Q(o, a)$. Concretely, we make $|r_e| \leq |\rlive|$. In our dense reward structure, an environment step either receives a positive reward of $\rdone$ and terminates the episode or receives a non-positive dense reward $r_\mathrm{dense}$. Thus, the highest $q$-value is achieved when a transition completes the task with reward $\rdone$. On the other hand, in the worst case where the agent receives $\rlive$ all the time and never succeeds, the $q$-value is bounded below by $\sum_{t=0}^\infty \gamma^t \rlive = (1 - \gamma)\rlive$. Thus, we apply the following clipping when computing the value target:
\begin{equation}
    (1 - \gamma)\rlive \leq Q(o, a) \leq \rdone.
\end{equation}
In essence, the above value clipping exploits our domain knowledge of the reward structure to bound the value function, which stabilizes learning.

\section{Experiments}
\subsection{Motivating Example}

\begin{figure}[t]
    \centering
    \begin{subfigure}[b]{2.7cm}
    \centering
    \includegraphics[height=2.5cm]{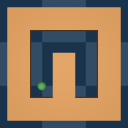}
    \caption{\small $s = (0.8, 0.8)$}
    \end{subfigure}
    \begin{subfigure}[b]{2.7cm}
    \centering
    \includegraphics[height=2.5cm]{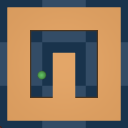}
    \caption{\small $s = (0.8, 1.3)$}
    \end{subfigure}
    \begin{subfigure}[b]{2.7cm}
    \centering
    \includegraphics[height=2.5cm]{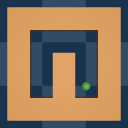}
    \caption{\small $s = (2.8, 0.8)$}
    \end{subfigure}
    \caption{\small Observations from the PointMaze environment. The point-mass in green is the controllable agent. $s$ is the agent's location. Clearly, states $a$ and $b$ are closer to each other than $a$ and $c$ are.}
    \label{fig:point_maze}
\end{figure}

We first motivate the need for learning a latent representation of a state and using our augmentation-invariant distance metric. In this experiment, we show that it is non-trivial to find a good distance metric between states from their respective high-dimensional image observations. To provide a clear illustration, we use the PointMaze environment provided in the D4RL benchmark \cite{d4rl}. In this environment, the controllable point-mass agent is marked in green. In Figure \ref{fig:point_maze}, we place the point-mass at three different positions in a same U-shaped maze. From the ground-truth state information, we know that state (a) is much closer to state (b) than state (c).

\begin{figure*}[t]
    \centering
    \includegraphics[width=0.245\linewidth]{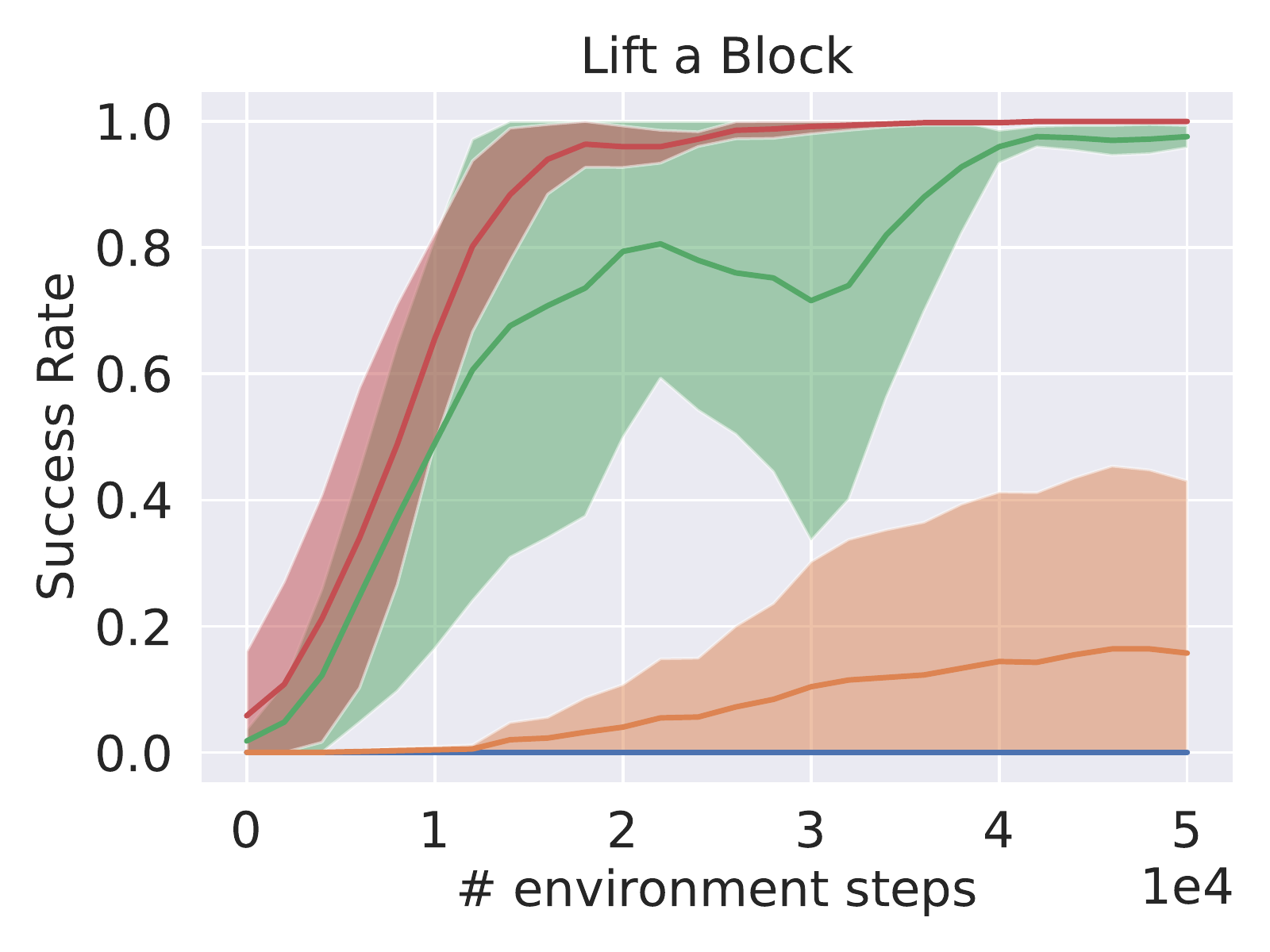}
    \includegraphics[width=0.245\linewidth]{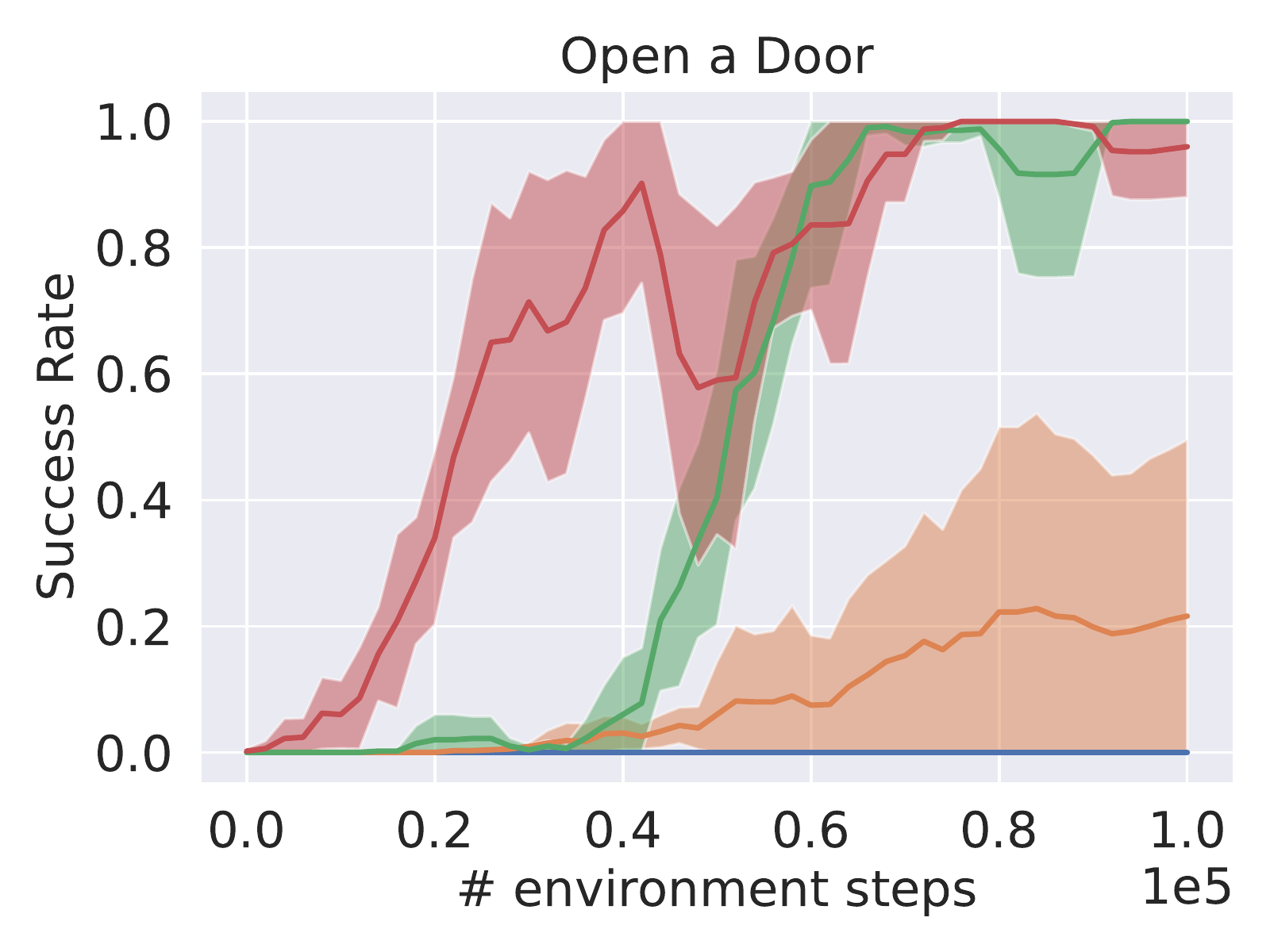}
    \includegraphics[width=0.245\linewidth]{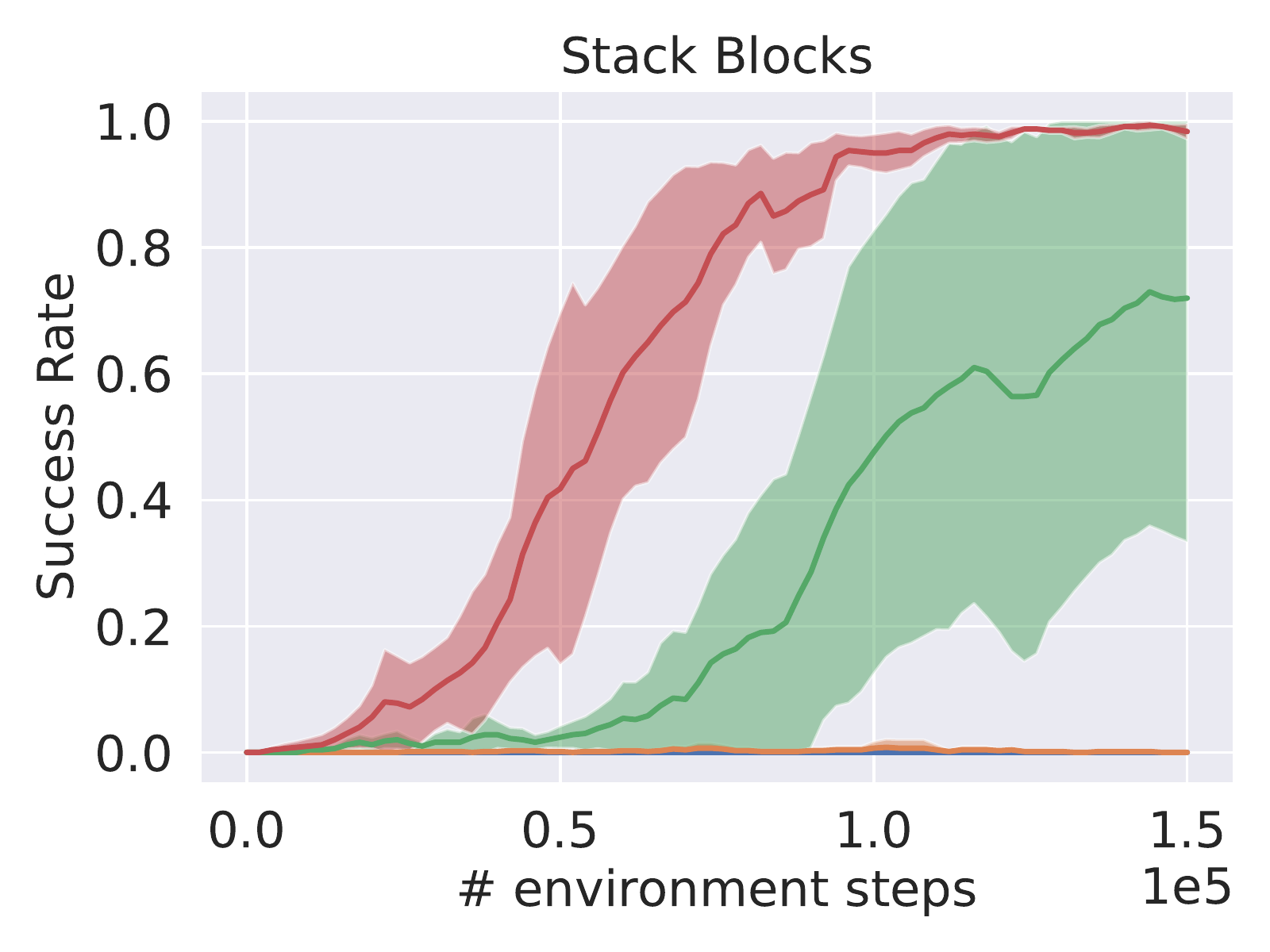}
    \includegraphics[width=0.245\linewidth]{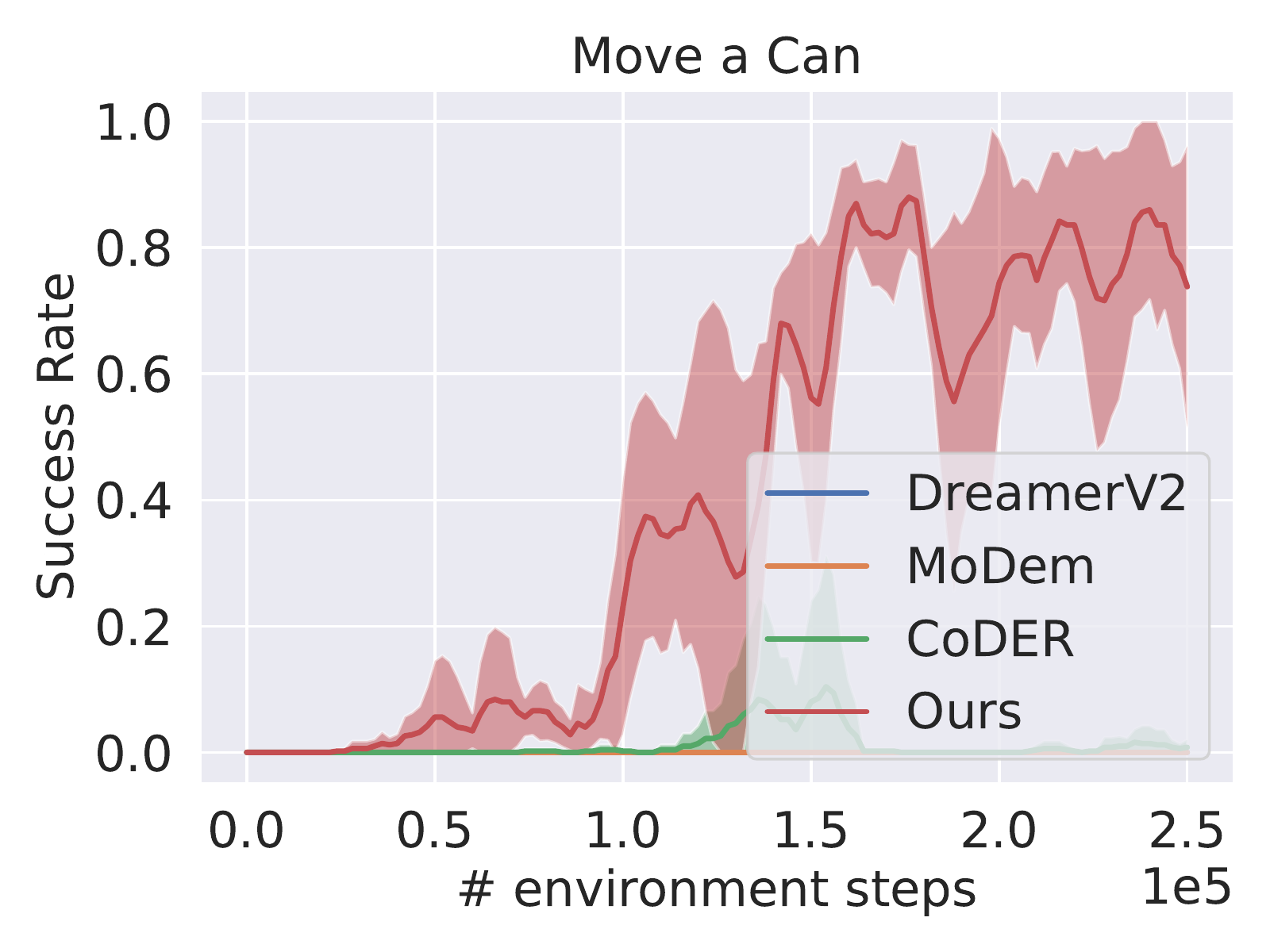}
    \caption{\small We compare our method with state-of-the art model-based and model-free methods, all given the same number of demonstrations. Clearly, our method (red) learns faster with a higher success rate than all three baseline methods.}
    \label{fig:results}
\end{figure*}

\begin{figure*}[t]
    \centering
    \includegraphics[width=0.245\linewidth]{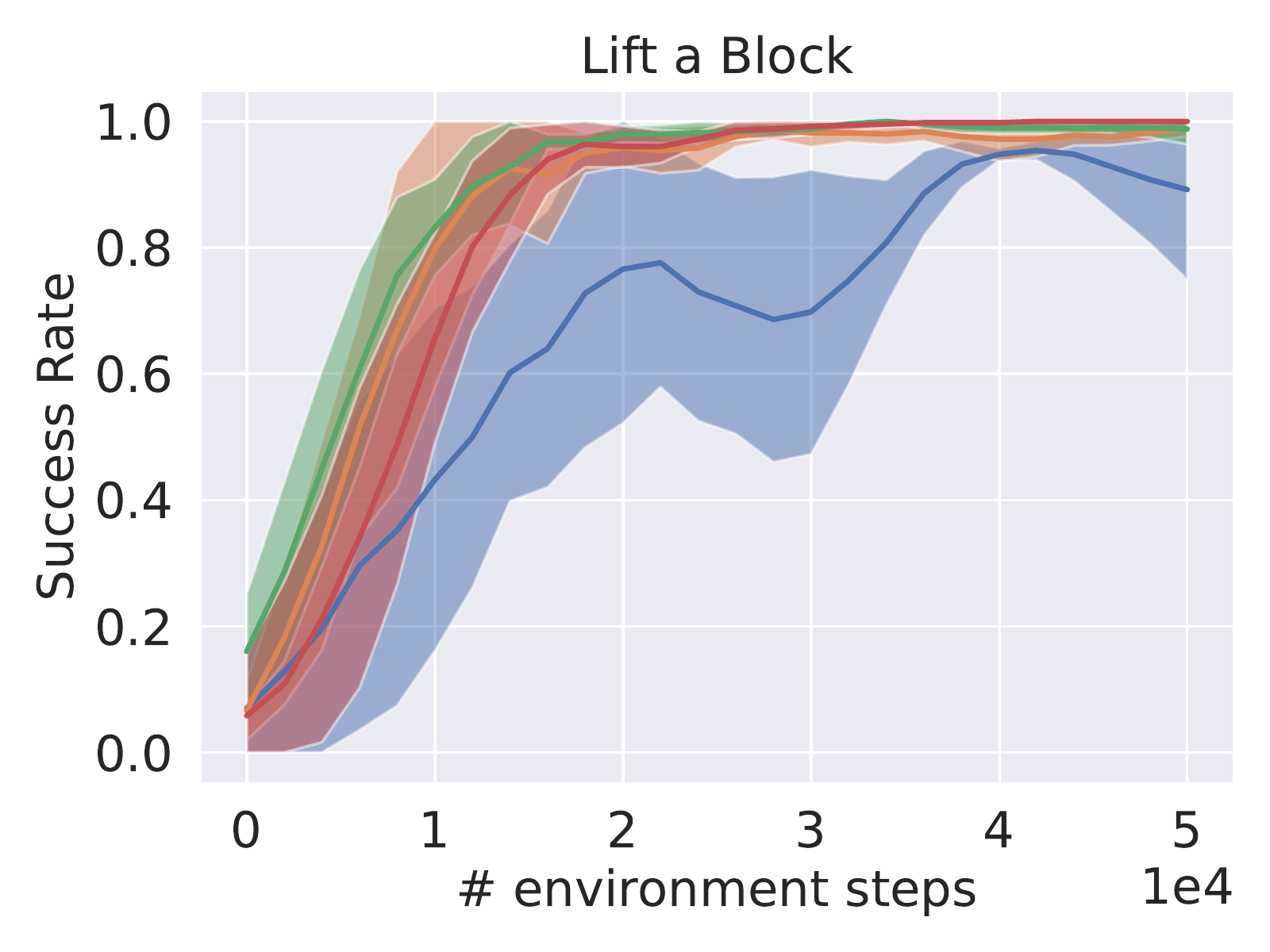}
    \includegraphics[width=0.245\linewidth]{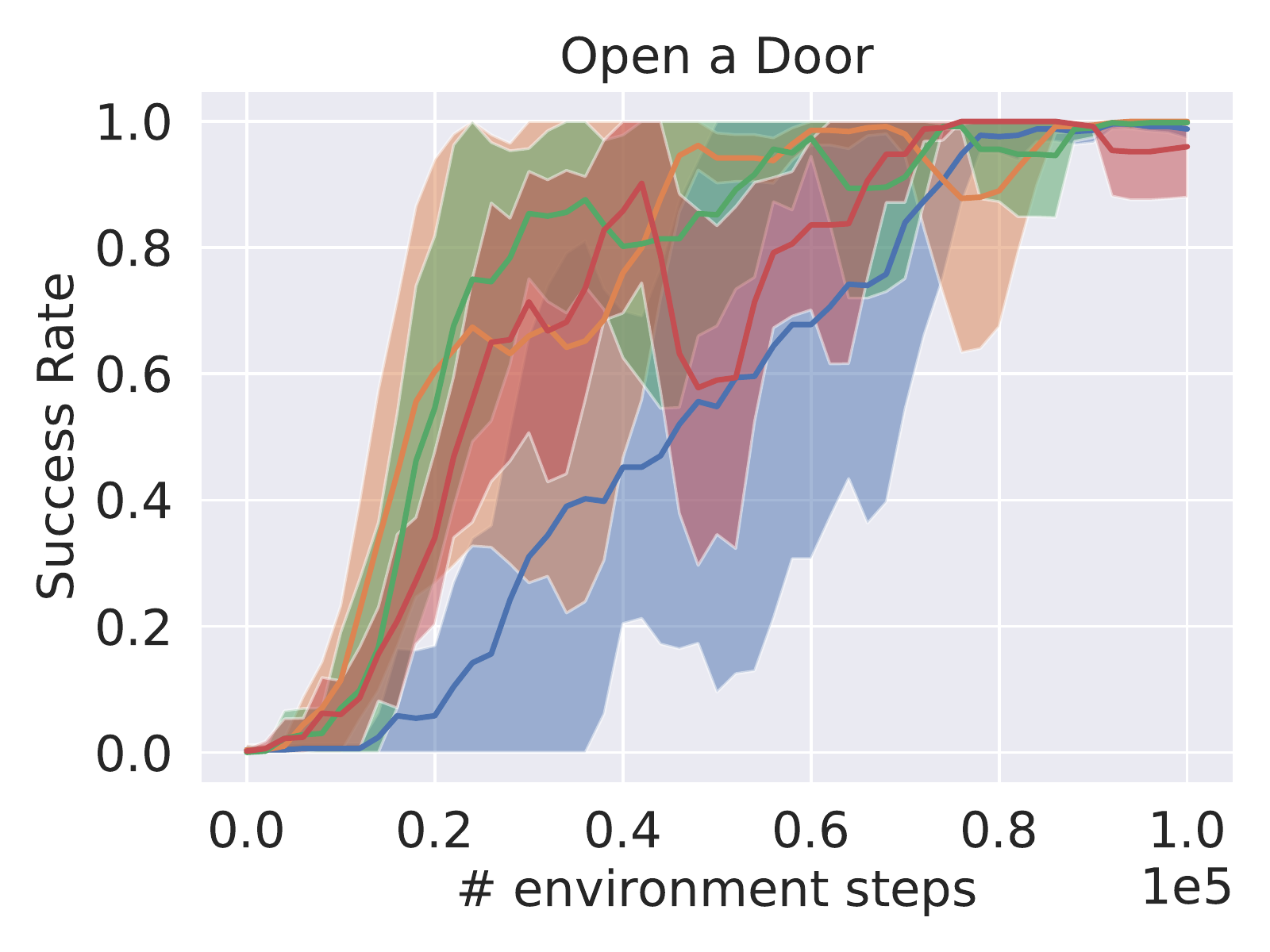}
    \includegraphics[width=0.245\linewidth]{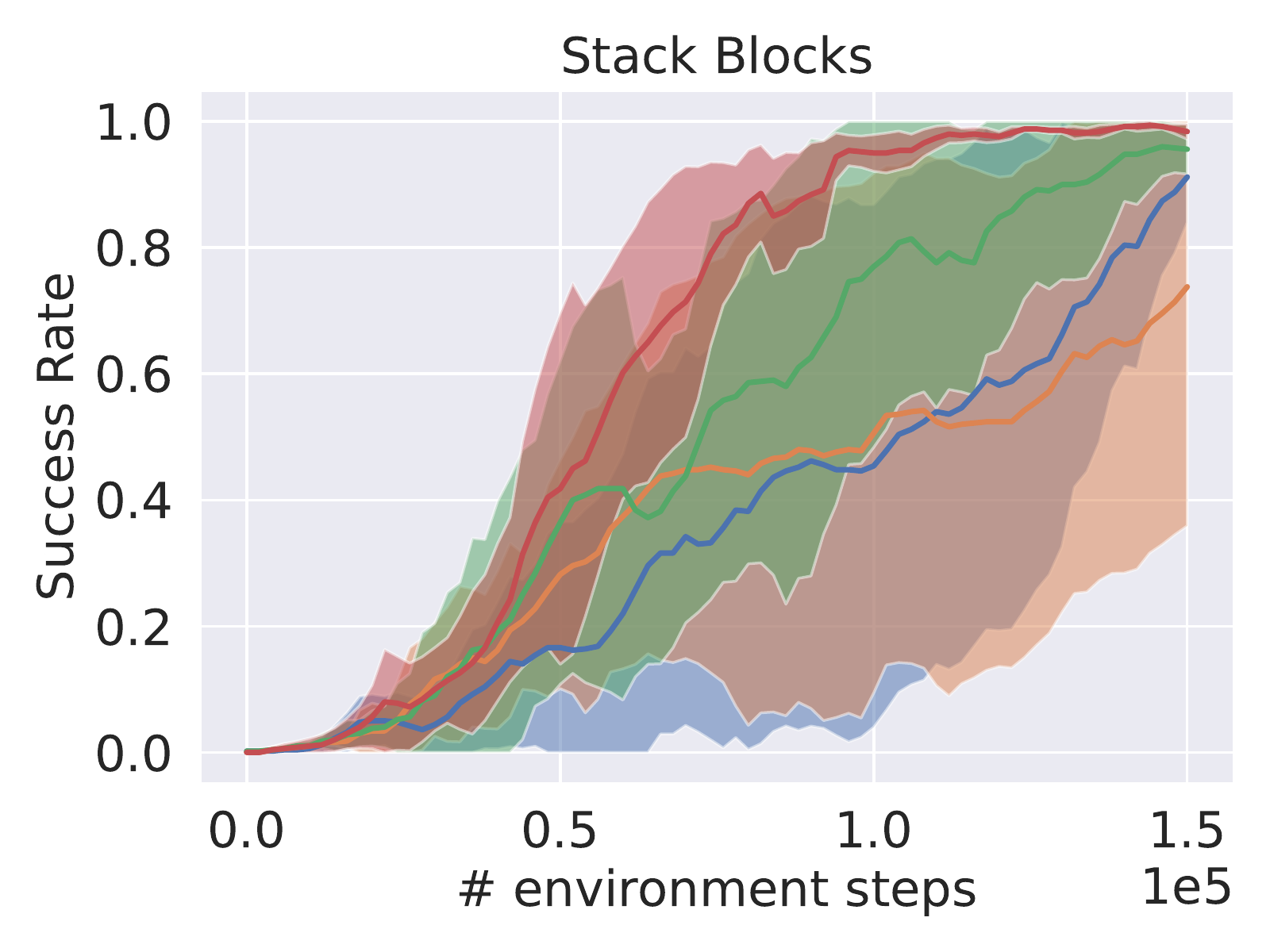}
    \includegraphics[width=0.245\linewidth]{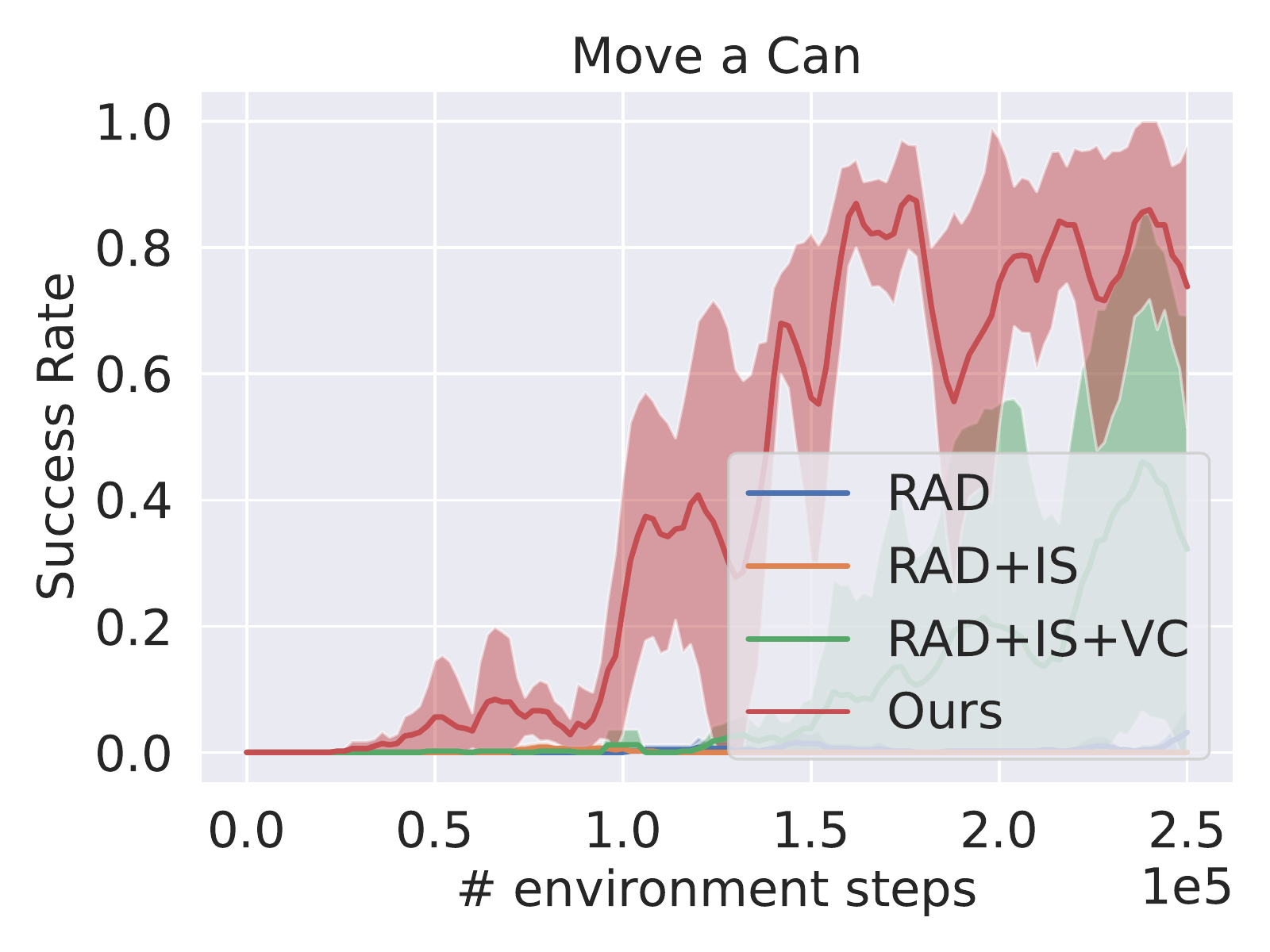}
    \caption{\small Starting from a standard algorithm (RAD), we successively add key improvements, including IS (importance sampling), VC (value clipping) and exploration reward. We observe that both importance sampling and value clipping are helpful in terms of propagating the training signal and stabilizing training. Finally, the additional exploration reward slows down training at the very beginning, but is crucial in the learning of long-horizon tasks.}
    \label{fig:ablation}
\end{figure*}

\begin{table}[h]
    \centering
    \begin{tabular}{| c | c | c | c |} 
      \hline
      Distance metric $d$ & $d(a, b)$ & $d(a, c)$ & $d(a, b) / d(a, c)$\\ 
      \hline
      Ground truth $L^2$ & 0.5 & 2.0 & \textbf{0.25} \\ 
      \hline
      Pixel $L^2$ & 4.102 & 4.178 & 0.982\\
      \hline
      VAE Embedding $L^2$ & 1.80e-1 & 1.53e-2 & 11.76\\
      \hline
      ADM & 1.092 & 2.255 & \textbf{0.484}\\
      \hline
    \end{tabular}
    \caption{\small Comparison across different distance metrics in PointMaze. Distance in the pixel space and VAE embedding space is unable to capture the true underlying states' relations. Whereas our ADM is able to estimate the relative distance most accurately.}
    \label{tab:point_maze}
\end{table}

In Table \ref{tab:point_maze}, we compare two baseline distance metrics with our proposed method ADM. The first row is the ground truth, indicating that indeed states $a$ and $b$ are closer than states $a$ and $c$. For Pixel $L^2$, we consider each image as a long vector, and directly compute the $L^2$ distance between them. For VAE Embedding $L^2$, we train a Variational Auto-encoder to reconstruct the image observations without considering the dynamics, and compute the distance in the embedding space. For a consistent comparison, we don't apply data augmentations in this experiment. As expected, $d(a, b)$ and $d(a, c)$ are not distinguishable from the pixel-wise distance. While the VAE Embedding distance is able to reconstruct the images, it doesn't incorporate the transition information, and fails to estimate the distance between states. ADM is the only method to capture the relative proportions of the ground-truth distances. This simple experiment illustrates the benefits of using our latent space distance metric, ADM, to measure task-relevant similarity between image observations.

\subsection{Robot Manipulation from Sparse Rewards}

\begin{figure}[h]
    \centering
    \begin{subfigure}[b]{0.4\linewidth}
    \centering
    \includegraphics[width=0.9\linewidth]{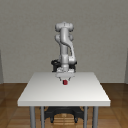}
    \caption{\small Lift a Block}
    \end{subfigure}
    \begin{subfigure}[b]{0.4\linewidth}
    \centering
    \includegraphics[width=0.9\linewidth]{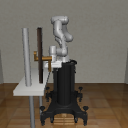}
    \caption{\small Open a Door}
    \end{subfigure}\\
    \vspace{0.2cm}
    \begin{subfigure}[b]{0.4\linewidth}
    \centering
    \includegraphics[width=0.9\linewidth]{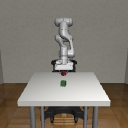}
    \caption{\small Stack Blocks}
    \end{subfigure}
    \begin{subfigure}[b]{0.4\linewidth}
    \centering
    \includegraphics[width=0.9\linewidth]{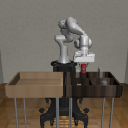}
    \caption{\small Move a Can}
    \end{subfigure}
    \caption{\small We experiment with 4 diverse manipulation tasks in the RoboSuite simulator. Shown above are observations from the front-view camera. The other camera is mounted on the robot's end effector. These 2 RGB images are the only observations provided to the RL agent.}
    \label{fig:robosuite}
\end{figure}

To verify the capability of our proposed approach, we aim to solve 4 robot manipulation tasks from the Robosuite simulator \cite{robosuite2020}, including picking up a block, stacking blocks, opening a door, and moving a soda can. In particular, moving the soda can requires a pick-and-place motion that takes around 100 steps to complete. The observations provided to the agent at each timestep are $128 \times 128$ RGB images from two cameras, one mounted on the wrist of the robot arm and one in the front. We use Operational Space Control to move the robot -- the RL agent outputs actions to change the displacement, rotation and opening of the robot hand, for a total of 7 degrees of freedom. For a better illustration, Figure \ref{fig:robosuite} shows the front-view renderings from all four environments.

We use a hard-coded policy to collect demonstrations. Importantly, the scripted policy utilizes ground-truth state information and renders the image observations alongside to save in the demonstration set. The state information is not available for the RL agents, which must learn from camera observations alone. The number of demonstration trajectories and total steps collected for each task are shown in Table \ref{tab:demo_samples}. We emphasize that our method requires only a few demonstrations which are quick to collect even on real robot platforms, as we show in the next section.

\begin{table}[h]
    \centering
    \begin{tabular}{| c | c | c | c | c |} 
      \hline
      Task & Lift & Open a Door & Stack & Move a Can \\ 
      \hline
      \# demo & 5 & 10 & 10 & 20\\ 
      \# total steps & 94 & 538 & 548 & 1783 \\
      \hline
    \end{tabular}
    \caption{\small We only need a small number of demonstrations for each task. Among the tasks, Moving a Can is the hardest as even the demonstrator requires about 90 steps to finish.}
    \label{tab:demo_samples}
\end{table}

\begin{figure*}[t]
    \centering
    \begin{subfigure}[b]{0.32\linewidth}
    \centering
    \includegraphics[width=0.9\linewidth]{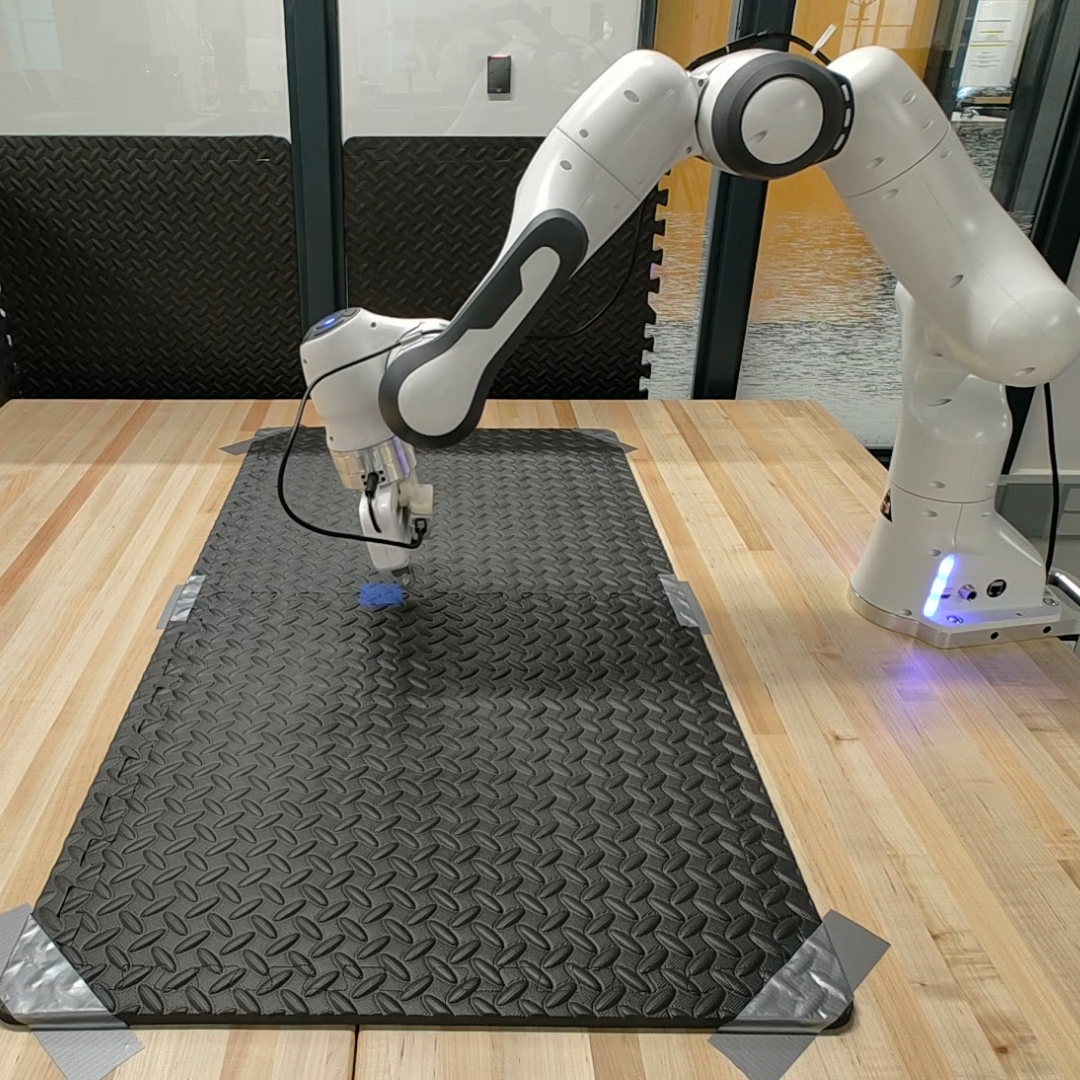}
    \caption{\small Reach a colored patch}
    \end{subfigure}
    \begin{subfigure}[b]{0.32\linewidth}
    \centering
    \includegraphics[width=0.9\linewidth]{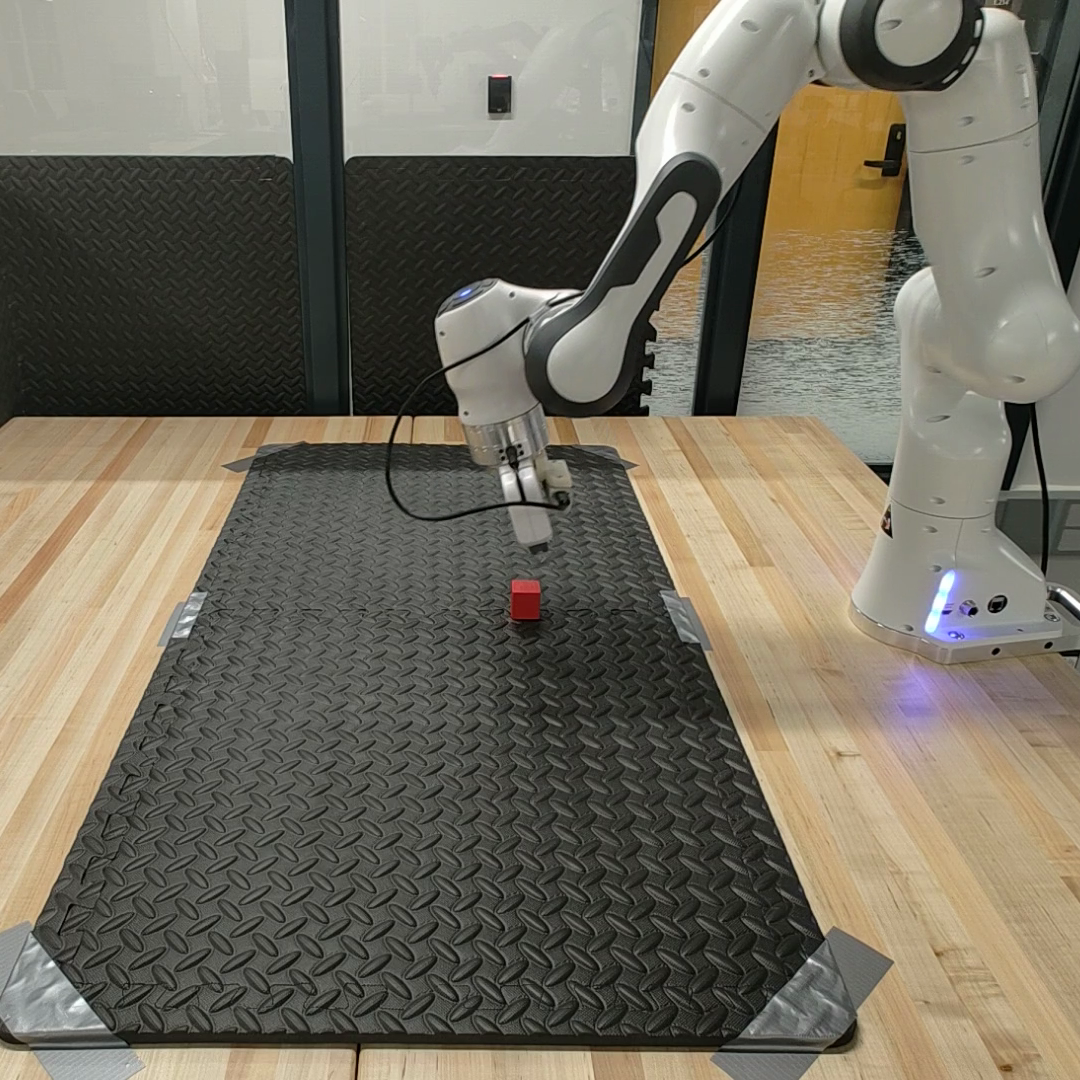}
    \caption{\small Reach a block}
    \end{subfigure}
    \begin{subfigure}[b]{0.32\linewidth}
    \centering
    \includegraphics[width=0.9\linewidth]{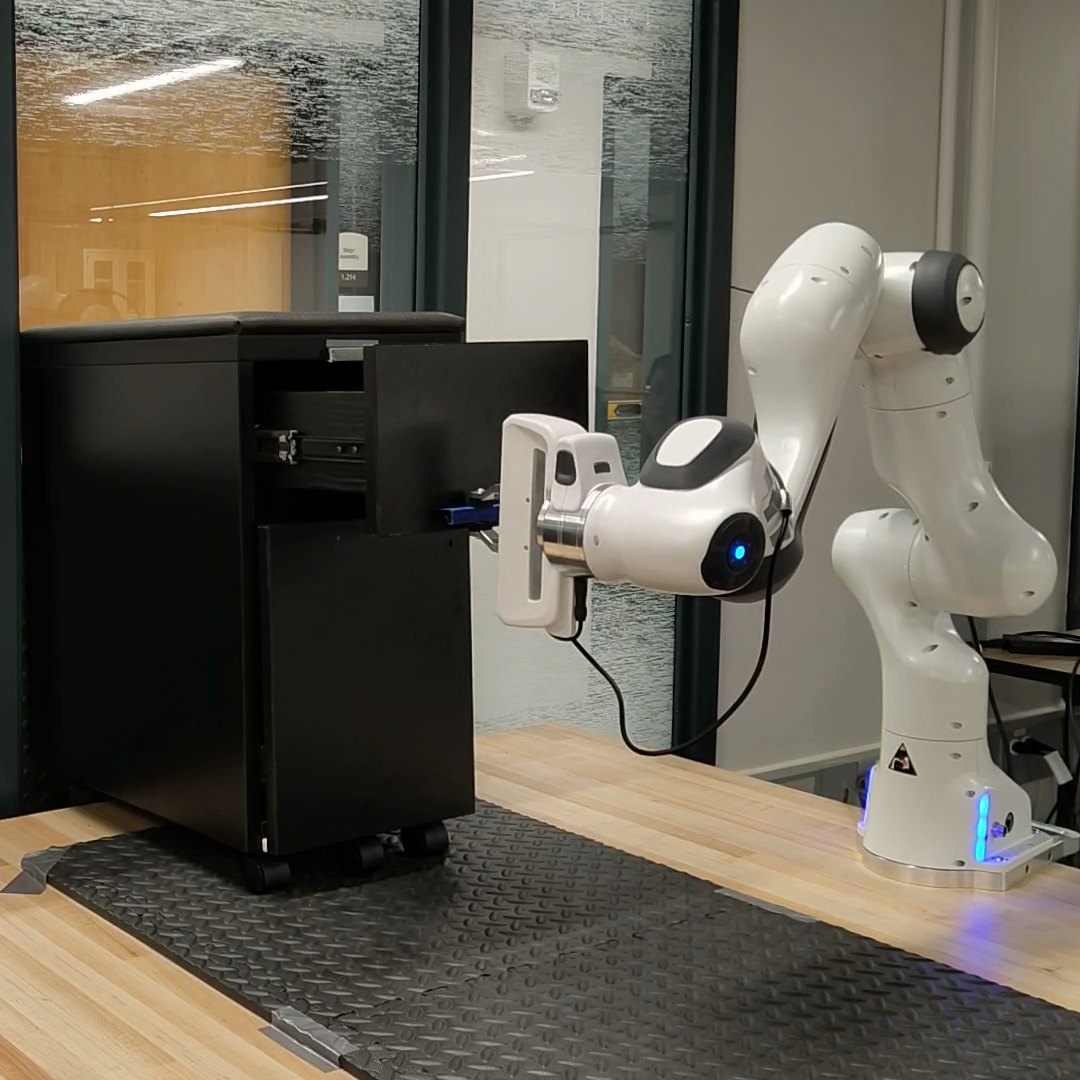}
    \caption{\small Open a drawer}
    \end{subfigure}
    \caption{\small We evaluate our method on 3 tasks with the Franka Emika Panda robot. The reinforcement learning agent controls the position and rotation of the robot's end effector, and additionally the gripper movement in task (b). The robot is able to achieve 100\% success rate on each task within a few hours in training.}
    \label{fig:franka_tasks}
\end{figure*}

We compare the performance of our approach with CoDER \cite{coder}, which is the state-of-the-art model-free approach for vision-based RL using a few demonstrations. Additionally, we compare with state-of-the-art model-based methods including MoDem \cite{modem} and DreamerV2 \cite{dreamerv2}. We initilize all methods with the same set of demonstrations. We measure the evaluation success rates during training, and repeat each experiment with 5 different random seeds to obtain the final results. In the case of MoDem and DreamerV2, to make our tasks compatible, we don't terminate the episode at success, but rather at a fixed number of steps. In Figure \ref{fig:results}, we report the mean and standard deviation of the evaluation success rates as a function of training environment steps. We observe that our method outperforms all three baselines across all four tasks, while showing major advantages in the two more challenging tasks. DreamerV2 does not explicitly take demonstrations. To make an apples-to-apples comparison with our work and other benchmarks, we provide demonstrations by initializing the training buffer with demonstrated episodes. However, DreamerV2 struggles to learn the tasks even with demonstrations as a result of the sparse reward structure.

\subsection{Ablation Studies}
Our approach utilizes multiple techniques to maximize the utility of demonstrations and to speed up learning. To gain more insight on how each component contributes to the overall performance, we perform ablation studies by starting from a standard RL algorithm RAD \citep{rad} (initialized with demonstrations) and introducing our key components one at a time, namely importance sampling (IS), value clipping (VC) and finally, our exploration reward $r_\mathrm{dense}$.

From the results shown in Figure \ref{fig:ablation}, we observe that both importance sampling and value clipping are crucial for our approach. In block-stacking, they result in faster convergence to the optimal policy, and in can-moving, they enable the agent to complete the task. Clearly, they are helpful in quickly propagating the reward signals from the demonstrations and stabilizing training of the $q$-values by utilizing prior knowledge on the structure of our sparse reward task. However, as can be seen from the hardest can-moving task, the exploration reward makes the most significant difference as it allows the robot to complete the task reliably much earlier into training. This is because it effectively matches to the closest demonstration and encourages meaningful exploration progress.

\section{Real Robot Training}
\label{sec:panda}

\subsection{Highly Compliant Real-time Controller}
Because reinforcement learning requires a trial-and-error process \cite{rl}, we expect the robot to make frequent physical contacts with the environment. In order to eliminate safety hazards, the controller must be compliant to external forces. On the other hand, we want to the robot to move swiftly so that each RL step takes less time to execute.

To this end, we develop a highly compliant real-time controller extended from the Cartesian Impedance controller \cite{impedance, jt}. At a high level, the end-effector tracks a equilibrium pose following a mass-spring-damper model. The robot asserts higher torque in the opposite direction as the current pose deviates further from the equilibrium. In the Cartesian space, we limit the maximum force exerted on the end effector by the robot, preventing it from causing damages. In the joint space, we apply a counter torque when a joint gets close to its hardware limit. The combination of these control rules builds a safety net around the robot for smoother RL training.

\subsection{System Architecture}
The robot gets sensory information from two Intel Realsense cameras -- one mounted on the end-effector and another on the side, as illustrated in Figure \ref{fig:franka}. While the cameras are capable of RGB-D images, our method only utilizes the color images. Both cameras are directly connected via USB to an Nvidia GPU desktop, which runs inference and training for the RL agent. Specifically, the GPU desktop runs an OpenAI gym interface \cite{gym}, to which the RL agent interacts with. At each timestep, the agent chooses its action based on its policy: $a = \pi(o)$. The action $a$ consists of a displacement of the end effector position, change in  roll/pitch/yaw, and open/close of the gripper. Next, the action selected by the RL agent is sent via ethernet to an Intel NUC, which directly interfaces with the Panda Robot. Specifically, the Intel NUC runs the Robot Operating System (ROS), where our real-time controller communicates with the \textsc{franka\_ros} library.

 \paragraph*{Collecting Demonstrations via Tele-Operation}
A key aspect of our approach is to efficiently learn from a small set of human demonstrations. We built an application where a user can tele-operate the robot by moving and rotating an iPhone. Our iPhone application utilizes primitives from Apple's ARKit \cite{linowes2017augmented} to stream the position and orientation of the device. During demonstration collection, a python script translates the tracking data into gym actions, and executes them on the real robot, which updates its equilibrium pose to follow the iPhone. These demonstration trajectories are stored on the Nvidia GPU desktop and used during training.

\subsection{Tasks}

We test on three diverse tasks on the Panda arm, as shown in Figure \ref{fig:franka_tasks}. We further introduce the reward function and resetting scheme as follows. 

\textit{Goal Reaching:} The simplest task is goal reaching, where the Panda must reach a fixed target patch of blue color. The sparse reward $\rdone = 100$ is given only when more than 25\% of the pixels from the gripper camera are blue, indicating we have correctly hovered over the target patch. The reward is $\rlive = -1$ everywhere else. The robot returns to its initial configuration at the beginning of each new episode.

\textit{Block Reaching:} Similar to goal-reaching, the task is for the robot's end effector to reach a randomly-placed red block. This task is challenging, because (1) an episode is deemed success only if more than 65\% of the bottom half pixels of the gripper camera are red, in which case the gripper will be able to grasp the block. This requires highly precise movements close to the goal. And (2) the block is randomly placed after a successful episode, requiring the robot to follow it to different locations instead of memorizing a fixed spot.

\textit{Opening a Drawer:} Another challenging task is to open a drawer. The robot must rotate from its original configuration and exercise fine control over its roll/pitch/yaw to grasp the drawer handle and open it, making use of the entire 7-dimensional action space. The sparse reward is given only if the drawer has been opened and extended by at least 50\%. When an episode succeeds, the robot moves in a pre-set path to close the drawer for automatic resetting.

\subsection{Results}
Experimental results show that our method is extremely data-efficient, leading to task success with only a couple of hours of training on the Panda Arm. The reason is that we effectively learn from demonstrations using our nearest-neighbor approach, which allows us to extrapolate a dense reward for the task. 

\begin{table}[h]
    \centering
    \begin{tabular}{| c | c | c | c |} 
      \hline
      Task & Reach Goal & Reach Block & Open Drawer \\ 
      \hline
      \# demo trajectories & 5 & 10 & 5\\ 
      \# demo steps & 63 & 171 & 128 \\
      Wall clock time & $\sim$2:00 & $\sim$6:00 & $\sim$4:00 \\
      \hline
      \# steps first success & 460 & 253 & 1169 \\
      \# steps convergence & 628 & 2714 & 2122 \\
      Wall clock time & 25:35 & 1:40:10 & 2:32:16\\
      \hline
    \end{tabular}
    \caption{\small In our real robot experiment, only a small number of demonstrations are needed for each task. All three tasks achieve 100\% success rate within a few hours of training.}
    \label{tab:franka-results}
\end{table}

Table \ref{tab:franka-results} shows the number of demonstrations provided and training performance of the three tasks completed by the Panda Arm. The Panda Arm was able to learn the task with only 30 minutes of training time for the goal reaching task. Compared to goal-reaching, it only took a couple of hours longer, including evaluation, to master harder tasks of block reaching and opening a drawer, all starting from scratch. Moreover, we point out that the wall clock time that we report are results prior to any computational optimization: we stop the robot and RL updates when we train the latent dynamics model, which consume about half the time. We also evaluate for 10 episodes every 10 training episodes, which is also counted in the total time.

Overall, our results on a real Panda Arm show the efficacy of learning a latent dynamics model and using the latent representation to find similar observations in expert demonstrations. 

\section{Limitations}
\label{sec:limitations}
Our approach does not automatically determine the optimal number of trajectories, and diversity within, the set of expert demonstrations. Moreover, it does not determine which states are truly \textit{causal} and directly lead to high rewards and task success. Finding the nearest causal state using our latent dynamics model and distance metric might improve the speed of learning. Finally, in our real robot framework, self-collision is handled by the robot's firmware, resulting in occasional delays during RL training for error-handling. Utilizing a path planner and actively preventing collisions might make training smoother.

\section{Conclusion}
\label{sec:conclusion}

This paper presents a data-efficient RL algorithm to learn sparse reward robotic tasks from image observations only, by utilizing a few demonstrations, which outperforms state-of-the-art model-free and model-based benchmarks. First, we start with a small set of expert demonstration trajectories. Then, we try to match a robot's current state with the closest state in a successful demonstration, in order to credit the robot for making meaningful exploratory progress. Our key insight is that simply matching states based on visual similarity is problematic -- the underlying states might be similar, but the high-dimensional pixel observations can differ due to background variation and task-irrelevant features. On the other hand, two images observations that differ only by a few pixels might require multiple steps in the environment. Thus, our key innovation is to learn a latent dynamics model, which provides a temporally consistent, concise latent state representation for each scene. Then, we find the closest latent state in a set of expert demonstrations (using data augmentation to improve robust matching) and assign an extra reward depending on the estimated number of steps away from the goal. As such, we can effectively convert a sparse reward task to a task with dense proxy rewards, which improves learning efficiency dramatically. 

Our work lends itself to several exciting future directions. First, we can leverage recent advancements in causal RL and counterfactual analysis \cite{bareinboim2016causal, mesnard2020counterfactual, zhu2019causal} to determine the state in an expert demonstration that directly caused high reward and task success. This might improve our search for nearest-neighbor demonstration states and overall learning. Second, we can utilize the latent dynamics model to further improve sample efficiency with model-based RL methods. Finally, we can explore theoretical guarantees for simple linear settings to show whether our proxy dense reward is optimal.

% \clearpage
\bibliographystyle{plainnat}
\bibliography{references}

\clearpage
\appendix
\section{Appendix}
\subsection{Environment Details}
\textit{Maximum Episode Length:} For our simulated experiments in Robosuite, we set maximum episode lengths based on the difficulty of each task, as shown in Table \ref{tab:episode_length}. Note that these numbers are chosen to be slightly over the average steps taken by the demonstrator to complete each task, leaving the RL agent plenty of time to finish. For our real robot experiments, all 3 tasks share the same maximum episode length of 30 steps. The episodes are terminated if the maximum length is reached or when the task is completed.

\begin{table}[h]
    \centering
    \begin{tabular}{| c | c | c | c | c |} 
      \hline
      Task & Lift & Open a Door & Stack & Move a Can \\ 
      \hline
      \# Max Episode Length & 40 & 80 & 80 & 120\\
      \hline
    \end{tabular}
    \caption{\small We choose the maximum episode length of each task to be slightly over the average steps taken by the demonstrator, giving the RL agent ample time to finish.}
    \label{tab:episode_length}
\end{table}

\textit{Task-specific Settings:} the block-lifting, door-opening and block-stacking tasks are all taken directly from the Robosuite simulator. The move-a-can task is a single-object version of the pick-and-place task where only the soda can is included. For the move-a-can task, the base of the robot is placed at the middle of the two bins.

\subsection{Neural Network Architectures}
Our method consists of 6 components parameterized by neural-networks, namely: model encoder $E_\phi$, model decoder $D_\theta$, locally-linear dynamics model $M_{\psi}$, RL encoder $E_\mathrm{RL}$, Actor $\pi_\mathrm{RL}$ and Critic $Q_\mathrm{RL}$. The numbers shown below corresponds to our specific setting where the latent space has 16 dimensions, and the action space has 7 dimensions. Our locally linear dynamics model predicts a low-rank approximation of $16 \times 16$ matrix $A$ using two $16$-dimensional vectors $u$ and $v$, where $A = I + uv^T$.

\begin{lstlisting}[basicstyle=\ttfamily\footnotesize]
Model Encoder:
Input: 6x112x112 randomly cropped images
ReLU(LayerNorm(Conv2D(6, 32, kernel=3, stride=2)))
ReLU(LayerNorm(Conv2D(32, 32, kernel=3, stride=1)))
ReLU(LayerNorm(Conv2D(32, 32, kernel=3, stride=1)))
Flatten()
ReLU(Linear(out_features=32))
ReLU(Linear(out_features=32))
Linear(out_features=32)
Output: 16-dim mean + 16-dim log-std
\end{lstlisting}

\begin{lstlisting}[basicstyle=\ttfamily\footnotesize]
Model Decoder:
Input: 16-dim latent vectors
ReLU(Linear(out_features=128))
ReLU(Linear(out_features=128))
ReLU(Linear(out_features=32768))
Reshape into 128x16x16
Upsample into 128x32x32
ReLU(Conv2D(128, 128, kernel=3, stride=1, pad=1)))
Upsample into 128x64x64
ReLU(Conv2D(128, 128, kernel=3, stride=1, pad=1)))
Upsample into 128x128x128
Conv2D(128, 6, kernel=3, stride=1, pad=1))
Output: 6x128x128 images
\end{lstlisting}

\begin{lstlisting}[basicstyle=\ttfamily\footnotesize]
Locally-Linear Dynamics model:
Input: 16-dim latent vectors
ReLU(Linear(out_features=512))
ReLU(Linear(out_features=512))
ReLU(Linear(out_features=160))
Output: 16-dim vector u + 16-dim vector 
        + 16x7 matrix B + 16-dim offset c
\end{lstlisting}

\begin{lstlisting}[basicstyle=\ttfamily\footnotesize]
RL Encoder:
Input: 6x112x112 randomly cropped images
ReLU(LayerNorm(Conv2D(6, 32, kernel=3, stride=2)))
ReLU(LayerNorm(Conv2D(32, 32, kernel=3, stride=2)))
ReLU(LayerNorm(Conv2D(32, 32, kernel=3, stride=2)))
ReLU(LayerNorm(Conv2D(32, 32, kernel=3, stride=2)))
Flatten()
LayerNorm(Linear(out_features=32))
Output: 32-dim feature
\end{lstlisting}

\begin{lstlisting}[basicstyle=\ttfamily\footnotesize]
Actor:
Input: 32-dim feature
ReLU(Linear(out_features=1024))
ReLU(Linear(out_features=1024))
Linear(out_features=14)
Output: 7-dim mean + 7-dim log-std
\end{lstlisting}

\begin{lstlisting}[basicstyle=\ttfamily\footnotesize]
Critic:
Input: 32-dim feature + 7-dim action
ReLU(Linear(out_features=1024))
ReLU(Linear(out_features=1024))
Linear(out_features=1)
Output: Q-value
\end{lstlisting}

\subsection{Hyper-parameters and Training Scheduling}
For all our experiments, we use discounting factor $\gamma=0.99$, exploration reward discount $\alpha=0.98$. Actor and critic learning rates are $1e-3$. Latent dynamics model learning rate is $4e-3$. Batch size $B = 128$. We choose different fraction of demonstrations $p_d$ when we perform importance sampling, as shown in the table below.

\begin{table}[h]
    \centering
    \begin{tabular}{| c | c | c | c | c | c |} 
      \hline
      Task & Lift & Open a Door & Stack & Move a Can & Real Robot\\ 
      \hline
      $p_d$ & 0.15 & 0.15 & 0.15 & 0.2 & 0.2\\
      \hline
    \end{tabular}
    \caption{\small We vary the fraction $p_d$ of demonstration interactions within each batch depending on the task we are training.}
    \label{tab:p_d}
\end{table}

For simulated environments, we perform one Actor-Critic update for each environment steps taken. After loading the demonstrations and for every 5000 environment steps, we update the latent dynamics model until convergence.

For real robot experiments, we update the latent dynamics model for every 300 steps taken by the robot. Since each steps takes about 0.5s to execute on the real robot, we keep performing SAC updates while the robot is in motion. This results in about 10 updates per environment step.

\end{document}